\documentclass{article}
\usepackage[preprint]{spconf}
\usepackage{amsmath,graphicx,hyperref,amssymb}
\usepackage{pifont,enumitem}
\newcommand{\cmark}{\ding{51}} 
\newcommand{\xmark}{\ding{55}} 

\copyrightnotice{\parbox{\textwidth}{\scriptsize
\copyright~2026 IEEE. Personal use of this material is permitted. Permission from IEEE must be obtained for all other uses, in any current or future media, including reprinting/republishing this material for advertising or promotional purposes, creating new collective works, for resale or redistribution to servers or lists, or reuse of any copyrighted component of this work in other works.}}

\title{A Dual-Modulation Framework for RGB-T Crowd Counting via Spatially Modulated Attention and Adaptive Fusion}

\name{Yuhong Feng\ \ \ Hongtao Chen\ \ \ Qi Zhang\sthanks{Corresponding author.}\ \ \ Jie Chen\ \ \ Zhaoxi He\ \ \ Mingzhe Liu\ \ \ Jianghai Liao}
\address{College of Computer Science
 and Software Engineering, Shenzhen University, Shenzhen, China}

\begin{document}
\ninept
\maketitle
\begin{abstract}
Accurate RGB-Thermal (RGB-T) crowd counting is crucial for public safety in challenging conditions. While recent Transformer-based methods excel at capturing global context, their inherent lack of spatial inductive bias causes attention to spread to irrelevant background regions, compromising crowd localization precision. Furthermore, effectively bridging the gap between these distinct modalities remains a major hurdle. To tackle this, we propose the Dual Modulation Framework, comprising two modules: Spatially Modulated Attention (SMA), which improves crowd localization by using a learnable Spatial Decay Mask to penalize attention between distant tokens and prevent focus from spreading to the background; and Adaptive Fusion Modulation (AFM), which implements a dynamic gating mechanism to prioritize the most reliable modality for adaptive cross-modal fusion. Extensive experiments on RGB-T crowd counting datasets demonstrate the superior performance of our method compared to previous works. Code available at https://github.com/Cht2924/RGBT-Crowd-Counting.
\end{abstract}
\section{Introduction}
\label{sec:intro}
Accurate crowd counting is essential for public safety and urban management. However, conventional methods relying solely on RGB image often degrade in performance under adverse conditions. To address this, RGB-Thermal (RGB-T) approaches improve performance by leveraging the complementarity of visual and thermal data. However, how to achieve precise spatial localization and effective cross-modal fusion remain the core challenges.

\begin{figure}[htb]
  \centering
  \includegraphics[width=\linewidth]{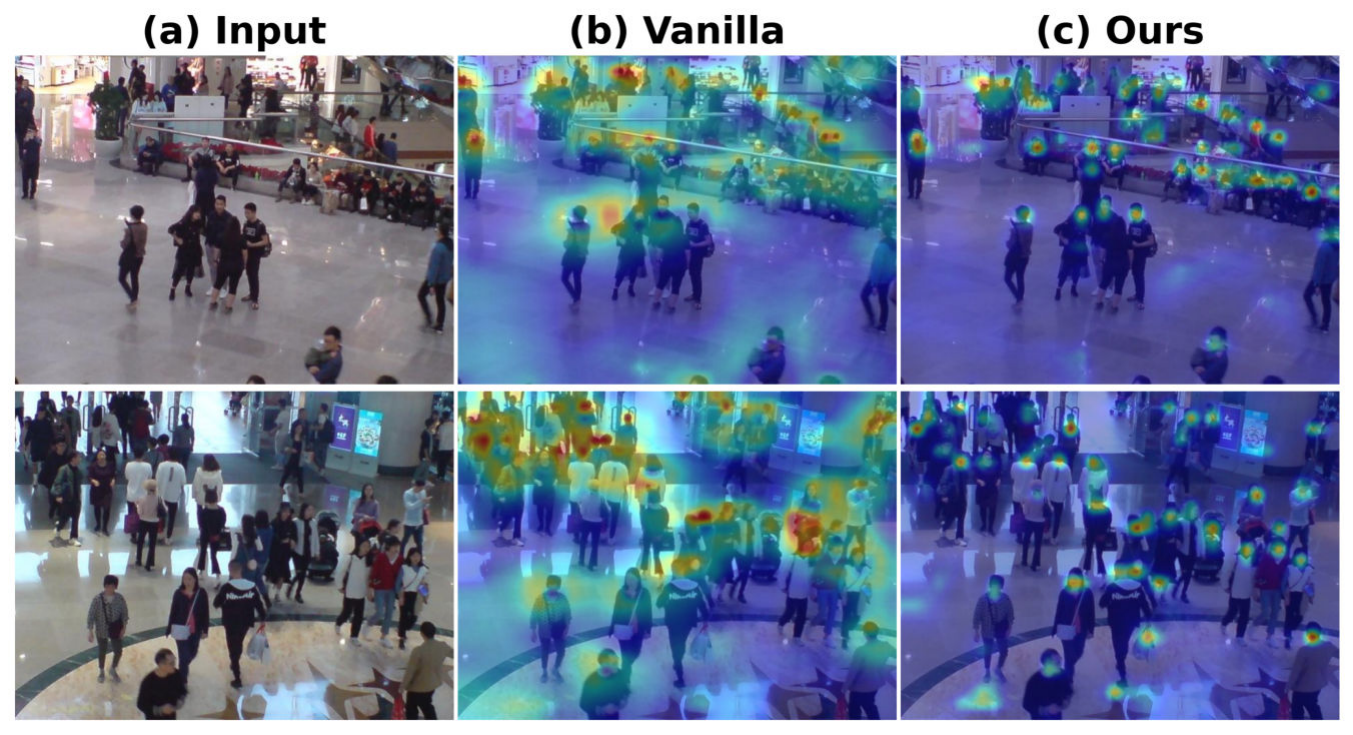}
  \caption{Our SMA prevents attention from spreading to irrelevant regions, enabling precise localization. (a) Input scene. (b) In the vanilla Transformer, attention incorrectly spreads to the background. (c) In contrast, our SMA produces compact, head-centric attention.}
  \label{fig:attn_vis}
\end{figure}

On the one hand, Transformers have emerged as leading architectures in multi-modal crowd counting, due to their ability to capture global contextual dependencies via self-attention mechanisms~\cite{liu2022rgbt, kong2024cross, wu2022multimodal, qu2024cascade, zhang2023cross, chang4984292dual, cai2023crowdfusion}. Despite their success, Transformers exhibit a fundamental architectural limitation in dense prediction tasks: their permutation-invariant design inherently lacks inductive biases for 2D spatial structure. As a result, the self-attention mechanism distributes focus across irrelevant background regions, as visualized in Fig.~\ref{fig:attn_vis}, which adversely affects density map estimation by leading to inaccurate localization and blurred boundaries. This deficiency severely impairs the model's ability to localize individual heads, particularly in dense crowd scenes. Some recent methods attempt to mitigate this by introducing structural priors. Gramformer~\cite{lin2024gramformer} tackles attention homogenization by building a graph based on feature dissimilarity to diversify attention maps, but it relies on computationally expensive graph operations.

On the other hand, inadequate cross-modal fusion remains a significant hurdle. The field has progressed to sophisticated multi-stage architectures~\cite{zhou2022defnet, meng2024multi}. These approaches leverage diverse strategies, from fixed structural priors~\cite{pan2024graph} to cross-modal attention~\cite{wang2024multi}. However, they exhibit a common limitation: their fusion logic focuses on the micro-level of feature interaction, lacking a macro-level, scene-aware regulatory strategy. This strategic deficiency makes them unable to dynamically adapt to varying environmental changes, like low light, and increase reliance on more informative thermal data when RGB inputs are compromised.

\begin{figure*}[h]
    \centering
    \includegraphics[width=\linewidth]{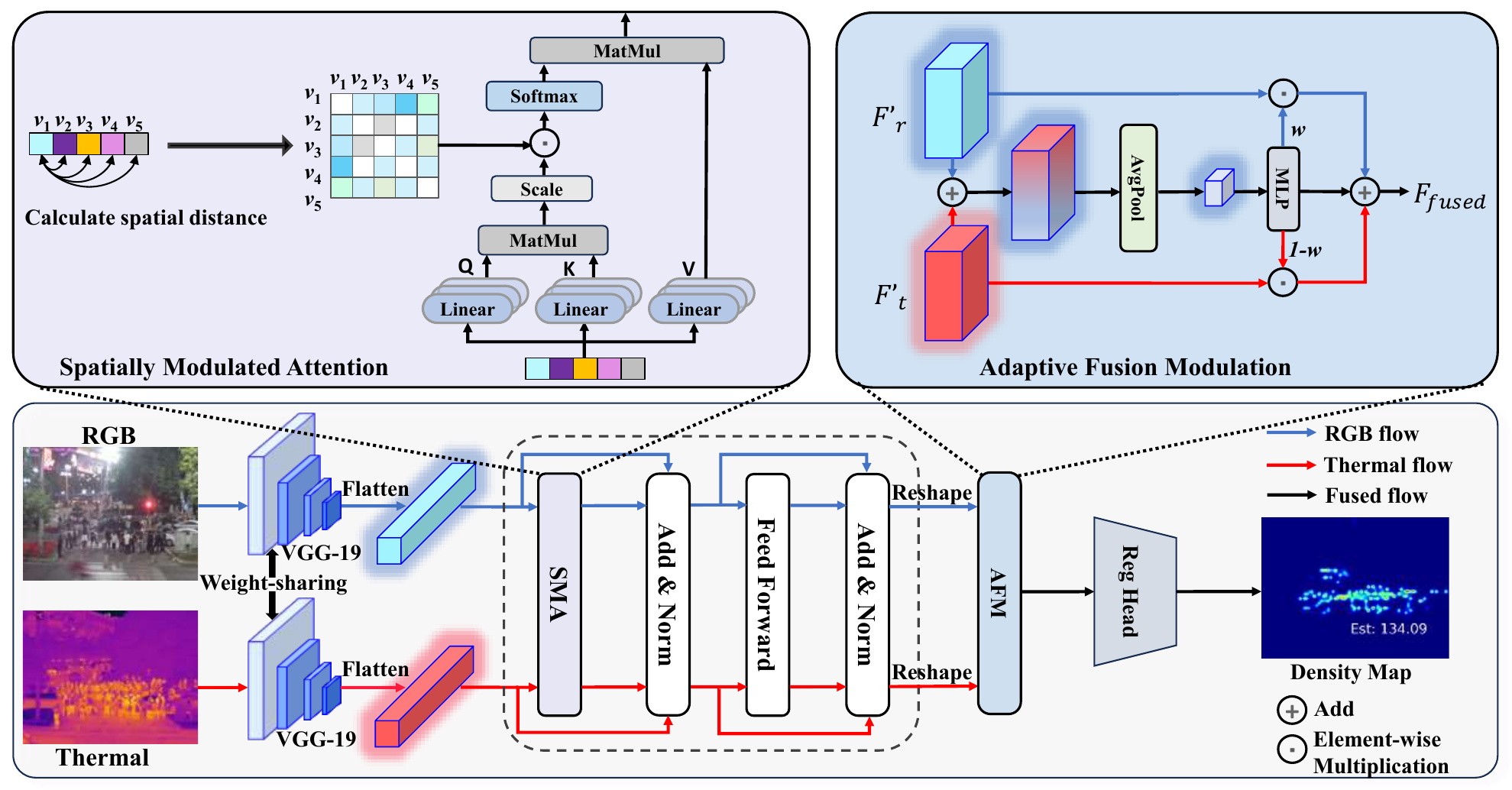}
    \caption{The architecture of our Dual Modulation framework. A shared VGG-19 backbone processes an RGB-T image pair, and the resulting features are processed by our SMA module, which improves spatial localization by enforcing a spatial inductive bias. The features are then fused by our AFM module, and a regression head produces the final density map.}
    \label{fig:model}
\vspace{-10pt}
\end{figure*}

To this end, we propose the Dual Modulation Framework, which is a unified framework designed to holistically address the dual challenges of achieving precise spatial localization and effective cross-modal fusion. First, to tackle the challenge of spatial localization, we propose the Spatially Modulated Attention (SMA) module. In contrast to modulating attention via complex, computationally expensive graph structures \cite{lin2024gramformer}, SMA offers a more direct and efficient solution. It injects a critical spatial inductive bias into the Transformer's self-attention mechanism by calculating the pairwise Euclidean distance between all tokens and applying a learnable Spatial Decay Mask. This process effectively suppresses spurious long-range token interactions and reduces background noise. Crucially, this learnable modulation empowers different attention heads to specialize: some learn a rapid decay to mimic local perception for fine-grained details, while others maintain a gentle decay to capture global context. This specialization effectively transforms the Transformer into a unified multi-scale encoder. Second, to address the challenge of cross-modal fusion, we propose the Adaptive Fusion Modulation (AFM) module. This module implements a lightweight, scene-aware fusion strategy at the macro level. It dynamically computes fusion weights for RGB and thermal features based on the scene content, enabling the model to intelligently adapt to environmental changes and prioritize the most reliable modality. The main contributions of this work are as follows:
\begin{itemize}[left=0em]
    \setlength{\itemsep}{0pt}
    \item We propose the SMA module, which injects a learnable spatial bias into Transformers. This prevents attention from spreading to background regions, significantly improving the precision of crowd localization.
    \item We propose an AFM module that implements a content-aware gating strategy to dynamically adjust the weights of each modality, ensuring effective cross-modal fusion across diverse conditions.
    \item We present a unified Dual Modulation Framework that integrates our proposed SMA and AFM modules. Extensive experiments conducted on two public RGB-T crowd counting datasets demonstrate that this framework achieves superior performance.
\end{itemize}

\section{METHODOLOGY}
\label{sec:format}
Our proposed Dual Modulation Framework is an end-to-end network designed to address the dual challenges of achieving precise spatial localization and effective cross-modal fusion in RGB-T crowd counting. As illustrated in Fig.~\ref{fig:model}, a weight-sharing VGG-19 backbone first extracts parallel feature maps, $F_r$ and $F_t$, from an RGB-T image pair. These features are then fed into separate Transformer encoders, where our SMA module enhances them with spatial awareness by injecting a learnable spatial inductive bias into the self-attention mechanism. The output feature maps from this stage are denoted as $F'_r$ and $F'_t$. Subsequently, our AFM module dynamically fuses the feature maps $F'_r$ and $F'_t$ into a single feature map, $F_{\text{fused}}$. Finally, a regression head converts this fused feature map into the final density map $D_{\text{est}}$.

\subsection{Spatially Modulated Attention}
\label{ssec:sma}
Standard self-attention lacks inductive biases for 2D spatial structures, causing attention to spread to irrelevant regions and undermining precise spatial localization. To address this, our approach is to introduce a learnable Spatial Decay Mask. Unlike the standard mechanism in which all tokens are effectively equidistant, our mask directly injects a explicit spatial inductive bias into the self-attention calculation. This bias is implemented by systematically penalizing attention scores between distant tokens based on their pairwise spatial distance, which forces the model to prioritize local interactions and suppresses interference from irrelevant background regions. The scaled dot-product attention is modified as follows:
\begin{equation}
    \text{Attention}(Q, K, V) = \text{softmax}\left(\frac{QK^T}{\sqrt{d_k}} \odot M\right)V,
\end{equation}
where $Q, K, V$ are the query, key, and value matrices, $d_k$ is the key dimension, and $M$ is our proposed Spatial Decay Mask.

The mask $M$ is derived from the pairwise Euclidean distance matrix $S \in \mathbf{R}^{N \times N}$ between all tokens. Instead of a fixed decay function, we introduce a flexible, learnable power-law decay. We define two learnable parameters per head: a scale $\beta_{\text{scale}}$ and a bias $\beta_{\text{bias}}$. The mask is computed as:
\begin{align}
    \beta'_{\text{scale}} &= \text{sigmoid}(\beta_{\text{scale}}) \qquad \beta'_{\text{bias}} = \text{softplus}(\beta_{\text{bias}}) \\
    S'_{ij} &= \text{LeakyReLU}(S_{ij} - \beta'_{\text{bias}}) \\
    M_{ij} &= (\beta'_{\text{scale}})^{S'_{ij}}.
\end{align}

This mechanism allows the network to learn an adaptive, multi-scale spatial bias. The learnable parameter $\beta'_{\text{bias}}$ defines a proximity threshold for each attention head. For token pairs with a distance below this threshold, the LeakyReLU output is near-zero, effectively preventing attention decay and preserving local interactions. Beyond this threshold, the decay activates, penalizing attention to distant tokens with a severity controlled by the learnable $\beta'_{\text{scale}}$.

\begin{figure}[htb]
  \vspace{-3pt}
  \centering
  \includegraphics[width=\linewidth]{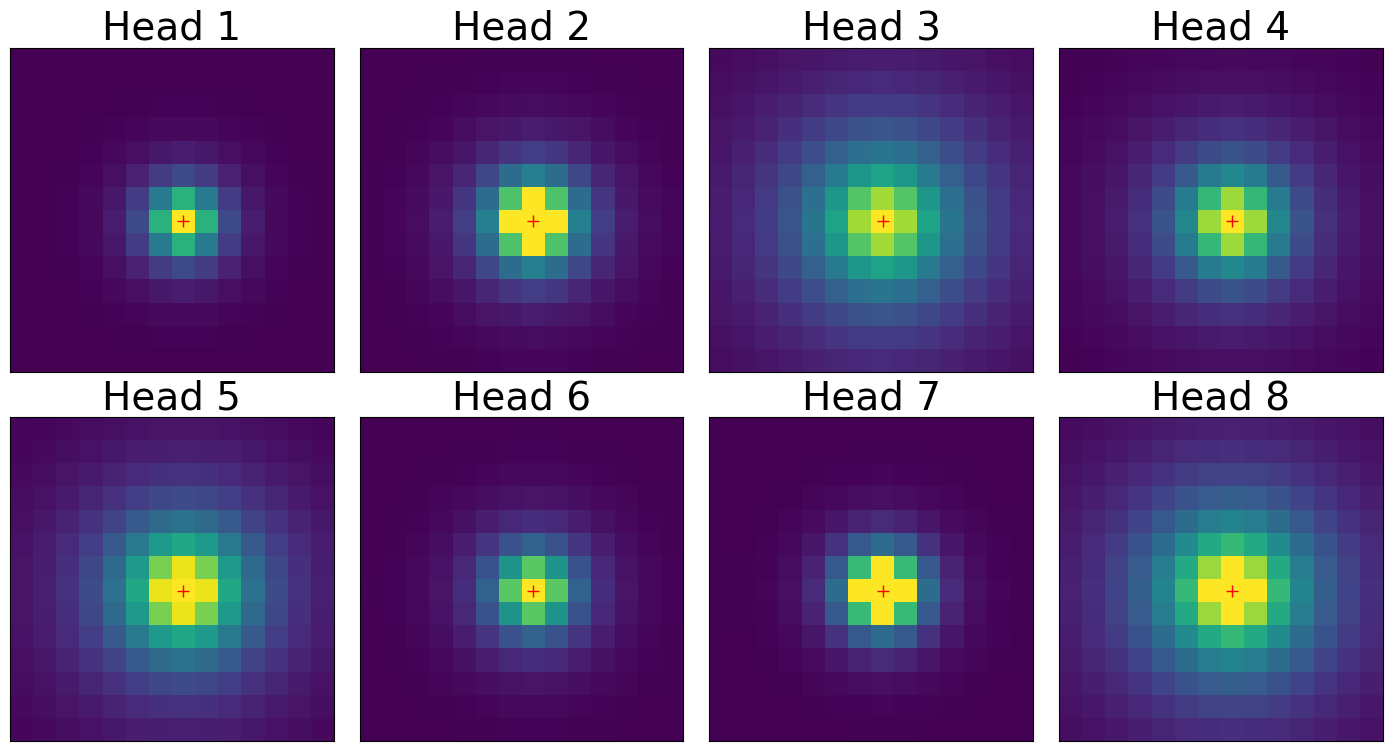}
  \caption{Visualization of the learned spatial decay patterns for each attention head. A clear specialization emerges: some heads learn tight, localized attention for fine-grained features. In contrast, others adopt a broader decay to capture global context.}
  \label{fig:decay}
\end{figure}

Crucially, each attention head learns a distinct spatial bias to specialize its function, effectively turning the Transformer into a unified multi-scale encoder. As visualized in Fig.~\ref{fig:decay}, this enables the capture of both local and global information. Some heads (e.g., Head 1 and 6) learn a rapid decay to mimic convolutional kernels for fine-grained features, while others (e.g., Head 3, 5, and 8) adopt a gentler decay to capture long-range context. This ability to process information across multiple scales in parallel empowers the model to capture rich multi-scale features, critical for accurate density estimation.

\vspace{-11pt}
\subsection{Adaptive Fusion Modulation}
\label{ssec:afm}
Effective fusion requires the model to dynamically assess the reliability of each modality under varying environmental conditions. To this end, we introduce the AFM module, a lightweight gating mechanism that learns to intelligently adjust the contribution of RGB and thermal features based on the holistic scene context.

Following the spatial modulation in the SMA, the resulting feature maps $F'_r$ and $F'_t$ are fed into the AFM module. We first create a joint feature map by element-wise summation: $F_{\text{sum}} = F'_r + F'_t$. This joint map, which encapsulates a summary of information from both modalities, is then globally pooled and passed through a lightweight MLP to generate a single fusion weight $w$. The MLP consists of two $1 \times 1$ convolutional layers with a final sigmoid activation:
\begin{align}
    w = \sigma(\text{Conv}_{2, 1 \times 1}(\text{ReLU}(\text{Conv}_{1, 1 \times 1}(\text{AvgPool}(F_{\text{sum}}))))),
\end{align}
where $\sigma$ is the sigmoid function. The resulting scalar weight $w$, a value between 0 and 1, represents a global, scene-level assessment of the relative reliability of the RGB modality. Consequently, $(1-w)$ represents the complementary importance of the thermal modality.

The final fused feature map $F_{\text{fused}}$ is computed via this adaptive weighted combination:
\begin{equation}
    F_{\text{fused}} = w \cdot F'_r + (1-w) \cdot F'_t.
\end{equation}

We opted for a global scalar weight $w$ over a spatially-varying weight map to enhance adaptability and prevent overfitting. This encourages the model to learn a global, scene-level fusion strategy rather than making noisy, pixel-wise decisions, which is more suitable for harmonizing features from different modalities. This mechanism allows the network to adapt to various conditions. For instance, in a low-light scene where RGB information is unreliable, the network learns to produce a small $w$, thereby prioritizing the more reliable thermal features to generate a more reliable final feature map.

\subsection{Regression Head and Loss Function}
\label{ssec:reg_loss}
Our regression head is composed of a stack of two $3 \times 3$ convolutional layers with intermediate ReLU activations. This is followed by a final $1 \times 1$ convolutional layer to produce the density map $D_{\text{est}}$.

Our network is trained with the Bayesian Loss~\cite{ma2019bayesian}, a common and effective choice for reconciling discrete point supervision with continuous density map regression. The loss is formulated as:
\begin{equation}
\mathcal{L}_{c} = \sum_{i=1}^{M} \left| 1 - \left\langle \frac{\mathcal{N}(z_i, \sigma^2 \mathbf{I}_{2\times2})}{\sum_{n=1}^{M} \mathcal{N}(z_n, \sigma^2 \mathbf{I}_{2\times2})}, D_{\text{est}} \right\rangle \right|.
\end{equation}
where $M$ signifies the total number of people in the image, $z_n$ indicates the coordinate of the $n$-th annotated head, and the $\langle \cdot, \cdot \rangle$ notation represents the inner product.

\section{EXPERIMENT}
\label{sec:experiments}

\begin{table*}[t]
\centering
\caption{Comparison with the state-of-the-art methods on RGBT-CC.}
\label{tab:rgbtcc}
\begin{tabular*}{\textwidth}{@{\extracolsep{\fill}}lcccccc}
\hline
Method & Venue & GAME(0) $\downarrow$ & GAME(1) $\downarrow$ & GAME(2) $\downarrow$ & GAME(3) $\downarrow$ & RMSE $\downarrow$ \\
\hline
MVMS~\cite{zhang2019wide} & CVPR 2019 & 19.97 & 25.10 & 31.02 & 38.91 & 33.97 \\
BBSNet~\cite{fan2020bbs} & ECCV 2020 & 19.56 & 25.07 & 31.25 & 39.24 & 32.48 \\
BL+IADM~\cite{liu2021cross} & CVPR 2021 & 15.61 & 19.95 & 24.69 & 32.89 & 28.18 \\
CSCA~\cite{zhang2022spatio} & ACCV 2022 & 14.32 & 18.91 & 23.81 & 32.47 & 26.01 \\
BL+MAT+SSP~\cite{wu2022multimodal} & ICME 2022 & 12.35 & 16.29 & 20.81 & 29.09 & 22.53 \\
MSDTrans~\cite{liu2022rgbt} & BMVC 2022 & 10.90 & 14.81 & 19.02 & 26.14 & 18.79 \\
MC$^3$Net~\cite{zhou2023mc} & TITS 2023 & 11.47 & 15.06 & 19.40 & 27.95 & 20.59 \\
GETANet~\cite{pan2024graph} & GRSL 2024 & 12.14 & 15.98 & 19.40 & 28.61 & 22.17 \\
MJPNet-T~\cite{zhou2024mjpnet} & IoT 2024 & 11.56 & 16.36 & 20.95 & 28.91 & \textbf{17.83} \\
MCN~\cite{kong2024cross} & ESWA 2024 & 11.56 & 15.92 & 20.16 & 28.06 & 19.02 \\
BGDFNet~\cite{xie2024bgdfnet} & TIM 2024 & 11.00 & 15.04 & 19.86 & 29.72 & 19.05 \\
VPMFNet~\cite{mu2024visual} & IoT 2024 & 10.99 & 15.17 & 20.07 & 28.03 & 19.61 \\
MISF-Net~\cite{gao2025three} & ISJ 2025 & 10.90 & 14.87 & 19.65 & 29.18 & 19.42 \\
\hline
Ours & - & \textbf{10.80} & \textbf{14.20} & \textbf{17.94} & \textbf{23.82} & 19.87 \\
\hline
\end{tabular*}
\vspace{-20pt}
\end{table*}

\textbf{Datasets.} Our experiments are conducted on two public RGBT crowd counting datasets.
RGBT-CC~\cite{liu2021cross} is a large-scale RGBT crowd counting benchmark, containing 2,030 RGB-T image pairs of size 640 × 480, with a total of 138,389 annotated pedestrians. The dataset is officially split into 1,030 pairs for training, 200 for validation, and 800 for testing.
DroneRGBT~\cite{peng2020rgb} is a large-scale, drone-based benchmark containing 3,607 RGB-T image pairs with a resolution of 640 × 512. The dataset is divided equally into 1,807 pairs for training and 1,800 for testing.

\begin{figure}[hb]
    \centering
    \includegraphics[width=\linewidth]{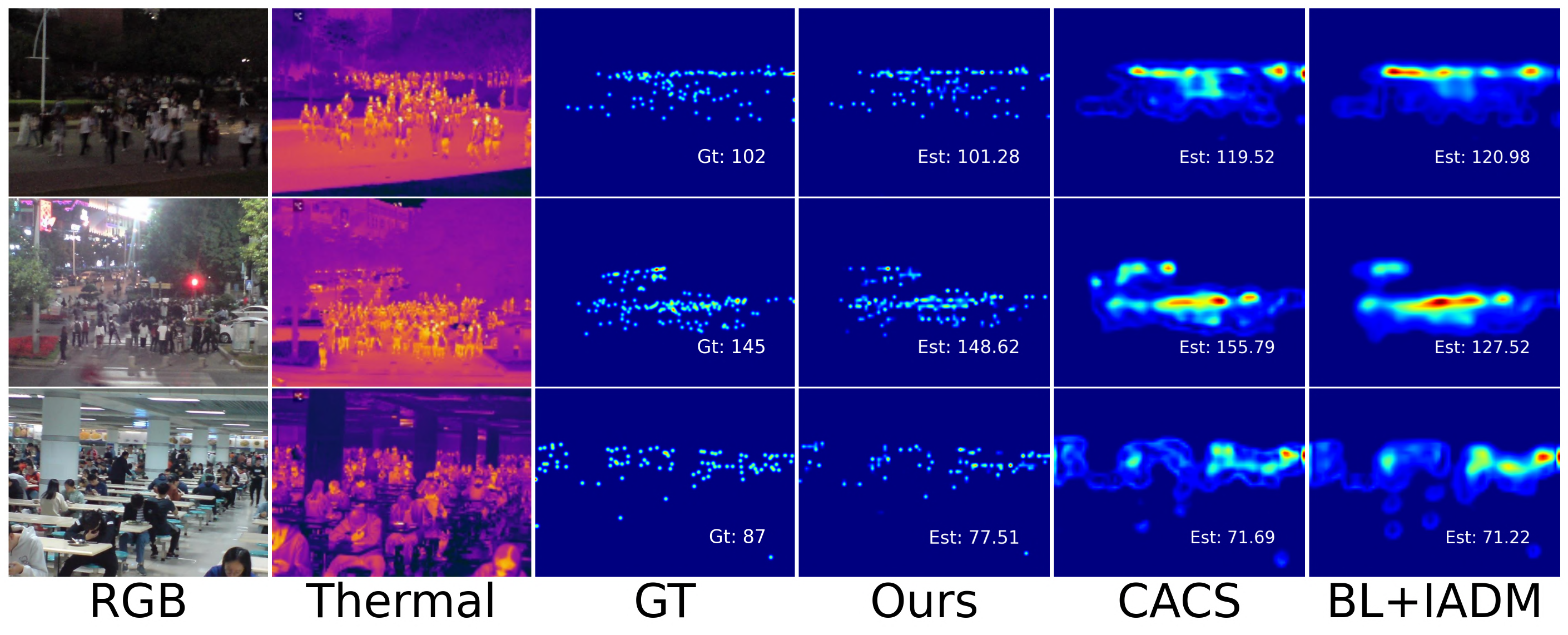}
    \caption{Visualization of crowd density maps generated by our model and other competing methods on several challenging scenes.}
    \label{fig:result}
\end{figure}

\noindent\textbf{Evaluation Metrics.} We follow standard practice and use Grid Average Mean absolute Error (GAME) and Root Mean Square Error (RMSE) as evaluation metrics. The GAME metric evaluates performance on subdivisions, where the image is divided into $4^L$ non-overlapping regions at level $L$. It is defined as: 
\begin{equation} \text{GAME}(L) = \frac{1}{N} \sum_{i=1}^{N} \sum_{j=1}^{4^L} |C_{et}^i(j) - C_{gt}^i(j)| \end{equation} where $N$ is the number of test images, and $C_{et}^i(j)$ and $C_{gt}^i(j)$ are the estimated and ground-truth counts for the $j$-th region of image $i$, respectively. For both metrics, lower values indicate better performance.

\noindent\textbf{Implementation Details.} Our model is implemented in PyTorch and trained on a single NVIDIA RTX 3090 GPU. We use a pre-trained VGG19 as the backbone, followed by a 2-layer transformer encoder with 8 attention heads. The model is trained for 400 epochs with a batch size of 1, using the Adam optimizer with a learning rate of 1e-5 and a weight decay of 1e-4. For data augmentation, we apply random horizontal flipping and random cropping to 224$\times$224.

\noindent\textbf{Results and Analysis.} As shown in Table~\ref{tab:rgbtcc}, our proposed method achieves highly competitive performance on the RGBT-CC dataset. Our model demonstrates comprehensive superiority across all GAME metrics, achieving leading scores of 10.80, 14.20, 17.94, and 23.82 for GAME(0) through GAME(3), respectively. On the RMSE metric, our model also achieves a highly competitive performance with a score of 19.87. This comprehensive lead on all GAME metrics is significant, as it confirms a more precise estimation of crowd spatial distribution, a direct benefit of our SMA module.

On the DroneRGBT dataset, as detailed in Table~\ref{tab:dronergbd}, our model surpasses all existing methods by achieving a score of 6.97 on the GAME(0) metric and 11.35 on the RMSE metric.

\begin{table}[t]
\centering
\caption{Comparison with the state-of-the-art methods on DroneRGBT.}
\label{tab:dronergbd}
\begin{tabular*}{\columnwidth}{@{\extracolsep{\fill}}lccc}
\hline
Method & Venue & GAME(0) $\downarrow$ & RMSE $\downarrow$ \\
\hline
CSRNet~\cite{li2018csrnet} & CVPR 2018 & 8.91 & 13.80 \\
MMCCN~\cite{peng2020rgb} & ACCV 2020 & 7.27 & 11.45 \\
BL+IADM~\cite{liu2021cross} & CVPR 2021 & 11.41 & 17.54 \\
DEFNet~\cite{zhou2022defnet} & TITS 2022 & 10.87 & 17.93 \\
CMDBIT~\cite{xie2023cross} & TCSVT 2023 & 11.50 & 21.27 \\
CGINet~\cite{pan2023cginet} & EAAI 2023 & 8.37 & 13.45 \\
GETANet~\cite{pan2024graph} & GRSL 2024 & 8.44 & 13.99 \\
\hline
Ours & - & \textbf{6.97} & \textbf{11.35} \\
\hline
\end{tabular*}
\vspace{-15pt}
\end{table}

The superior results on these two diverse datasets validate our framework's strong performance and generalization capability. The qualitative results in Fig.~\ref{fig:result} further showcase our model's ability to generate high-quality density maps that closely align with the ground truth, even in challenging scenes with varying crowd densities and lighting conditions.

\noindent\textbf{Ablation Study.} To validate the effectiveness of our proposed components, we conduct ablation experiments on the RGBT-CC dataset. The results are shown in Table~\ref{tab:ablation}. Our baseline model consists of the VGG-19 backbone, a standard Transformer encoder for each modality, and simple element-wise summation for fusion.

(1) Effectiveness of SMA. As shown in Table~\ref{tab:ablation}, integrating the SMA module alone significantly reduces GAME(0) from 12.51 to 11.09 and RMSE from 26.69 to 20.53. This demonstrates that injecting a spatial inductive bias via SMA effectively reduces background noise and enhances the precision of head localization.

(2) Effectiveness of AFM. Replacing simple summation with the AFM module further reduces GAME(0) to 10.80 and RMSE to 19.87. This improvement validates our dynamic fusion strategy, which enhances performance by prioritizing the more reliable modality based on scene content.

(3) Effectiveness of the Learnable Decay Mechanism. We conducted a parameter sensitivity analysis using a control variable method to compare our learnable decay against several fixed-parameter configurations. As shown in Table~\ref{tab:ablation_decay}, our learnable approach is consistently superior, a superiority that stems from its ability to let each attention head focus on a different spatial scale. As visualized in Fig.~\ref{fig:decay}, this allows some heads to learn localized attention for fine-grained details, while others focus on capturing global context. This multi-scale processing capability is the key to generating high-quality density maps.

\begin{table}[t]
\centering
\caption{The impact of different components of the model on RGBT-CC.}
\label{tab:ablation}
\setlength{\tabcolsep}{2pt}
\begin{tabular}{cc|ccccc}
\hline
SMA & AFM & GAME(0) & GAME(1) & GAME(2) & GAME(3) & RMSE \\
\hline
\xmark & \xmark & 12.51 & 15.59 & 19.13 & 24.87 & 26.69 \\
\cmark & \xmark & 11.09 & 14.62 & 18.58 & 24.48 & 20.53 \\
\cmark & \cmark & \textbf{10.80} & \textbf{14.20} & \textbf{17.94} & \textbf{23.82} & \textbf{19.87} \\
\hline
\end{tabular}
\vspace{-15pt}
\end{table}

\begin{table}[t]
\centering
\caption{Ablation Study on Decay Parameters on the RGBT-CC.}
\label{tab:ablation_decay}
\setlength{\tabcolsep}{2pt}
\begin{tabular}{cc|ccccc}
\hline
$\beta_{\text{scale}}$ & $\beta_{\text{bias}}$ & GAME(0) & GAME(1) & GAME(2) & GAME(3) & RMSE \\
\hline
0.10 & 0.10 & 11.42 & 14.38 & 18.11 & 23.91 & 20.56 \\
0.10 & 0.05 & 11.64 & 14.90 & 18.49 & 24.44 & 20.15 \\
0.20 & 0.10 & 11.50 & 14.72 & 18.34 & 24.27 & 20.82 \\
0.80 & 0.10 & 11.79 & 14.90 & 18.58 & 24.18 & 21.64 \\
\multicolumn{2}{c|}{Ours}   
& \textbf{10.80} & \textbf{14.20} & \textbf{17.94} & \textbf{23.82} & \textbf{19.87} \\
\hline
\end{tabular}
\vspace{-15pt}
\end{table}

\section{CONCLUSION}
\label{sec:conclusion}
This paper introduces the Dual Modulation Framework, an approach for RGB-T crowd counting based on a dual modulation of spatial attention and cross-modal fusion. The framework comprises two core modules: Spatially Modulated Attention to enable precise localization, and Adaptive Fusion Modulation to ensure effective cross-modal fusion. Through quantitative and qualitative experiments on two public RGB-T datasets, we demonstrate that our approach achieves competitive performance and high effectiveness for crowd counting.

\vfill\pagebreak

\bibliographystyle{IEEEbib}
\bibliography{strings,refs}

\end{document}